\documentclass{article}

\PassOptionsToPackage{numbers,compress}{natbib}
\usepackage[preprint]{neurips_2026}

\usepackage[utf8]{inputenc} 
\usepackage[T1]{fontenc}    
\usepackage{hyperref}       
\usepackage{url}            
\usepackage{booktabs}       
\usepackage{amsfonts}       
\usepackage{nicefrac}       
\usepackage{microtype}      
\usepackage{xcolor}         
\usepackage{graphicx}
\usepackage{capt-of}
\usepackage{adjustbox}
\usepackage{makecell}
\usepackage{multirow}
\usepackage{tabularx}
\usepackage{dsfont}
\usepackage{enumitem} 
\usepackage{amsmath}
\usepackage[table]{xcolor}

\title{VP-VLA: Visual Prompting as an Interface for Vision-Language-Action Models}

\author{%
    \textbf{Zixuan Wang$^{1*}$ \hspace{8pt} Yuxin Chen$^{1*}$ \hspace{8pt} Yuqi Liu$^{2*}$ \hspace{8pt} Jinhui Ye$^{1}$ \hspace{8pt} Pengguang Chen$^{3}$} \\
    \textbf{Changsheng Lu$^{1}$ \hspace{12pt} Shu Liu$^{3}$ \hspace{12pt} Bei Yu$^{2}$ \hspace{12pt} Jiaya Jia$^{1,3}$} \\
    \vspace{2pt} \\
    HKUST$^{1}$ \hspace{15pt} CUHK$^{2}$ \hspace{15pt} SmartMore$^{3}$ \\
    {\tt\small \url{https://github.com/JIA-Lab-research/VP-VLA}}
}

\begin{document}

\maketitle
\begin{center}
    \includegraphics[width=1.\linewidth]{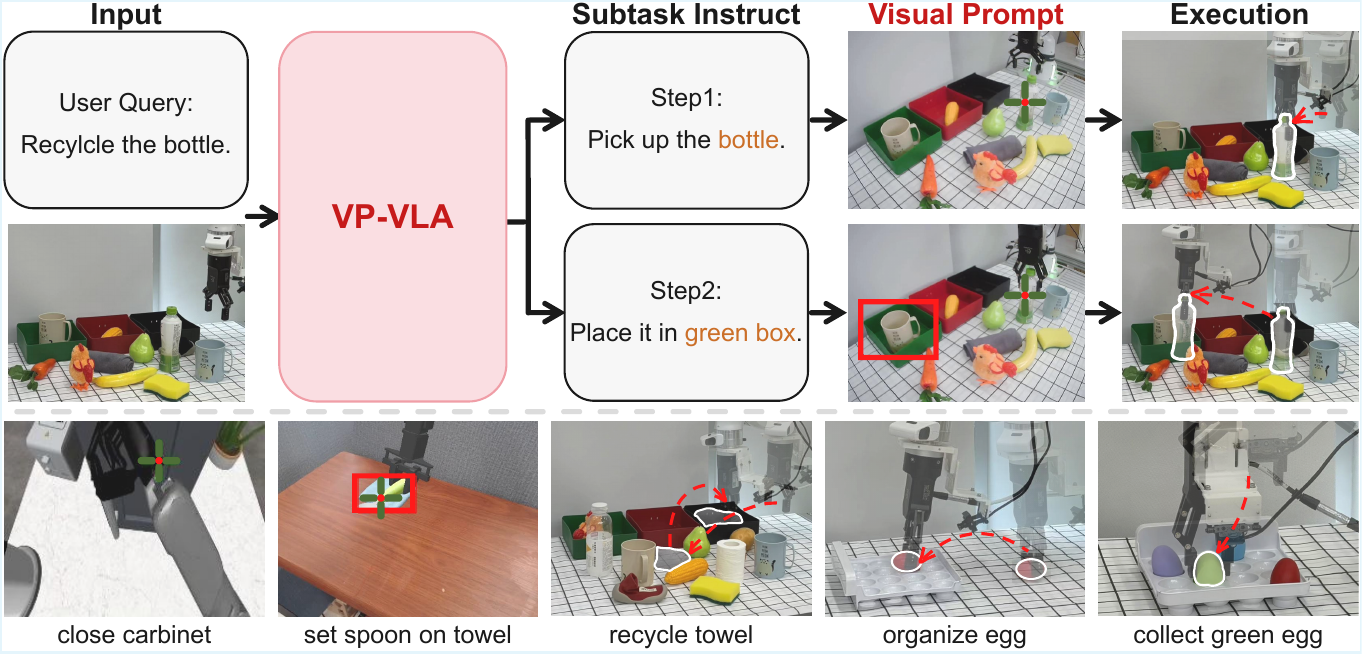}
    \captionof{figure}{VP-VLA leverages a dual-system architecture to bridge high-level reasoning and low-level control, maintaining competitive performance across a wide variety of tasks on in-distribution and out-of-distribution settings.}
    \label{fig:first_image}
\end{center}

\begin{abstract}
  Vision-Language-Action (VLA) models typically map visual observations and linguistic instructions directly to control signals. This ``black-box'' mapping forces a single forward pass to simultaneously handle instruction interpretation, spatial grounding, and low-level control, often leading to poor spatial precision and limited robustness in out-of-distribution scenarios. To address these limitations, we propose VP-VLA, a dual-system framework that decouples high-level reasoning and low-level execution via a structured visual prompting interface. Specifically, a ``System 2 Planner'' decomposes complex instructions into sub-tasks and identifies relevant target objects and goal locations. These spatial anchors are rendered directly within the native RGB observation space as modality-consistent visual prompts, such as crosshairs and bounding boxes. This avoids the modality mismatch introduced by dense masks, affordance maps, or additional control-specific representations. Guided by these prompts and enhanced by a novel auxiliary visual grounding objective during training, a ``System 1 Controller'' reliably generates precise low-level execution motions. Extensive experiments in simulation and real world demonstrate that VP-VLA surpasses state-of-the-art end-to-end baselines including QwenOFT and GR00T-N1.6.
\end{abstract}

\section{Introduction}
\label{sec:intro}
Recent advances in vision language models (VLMs) have revolutionized robotic manipulation. Vision-language-action (VLA) models, in particular, aim to bridge semantic understanding and low-level control by fine-tuning pretrained VLMs on large-scale robotic datasets. By doing so, these models inherit strong real-world priors while acquiring embodied skills, offering a promising path toward generalist manipulation policies~\cite{bjorck2025gr00t, black2024pi0, liu2024rdt}.

\begin{figure}[t]
    \centering
    \includegraphics[width=1.\linewidth]{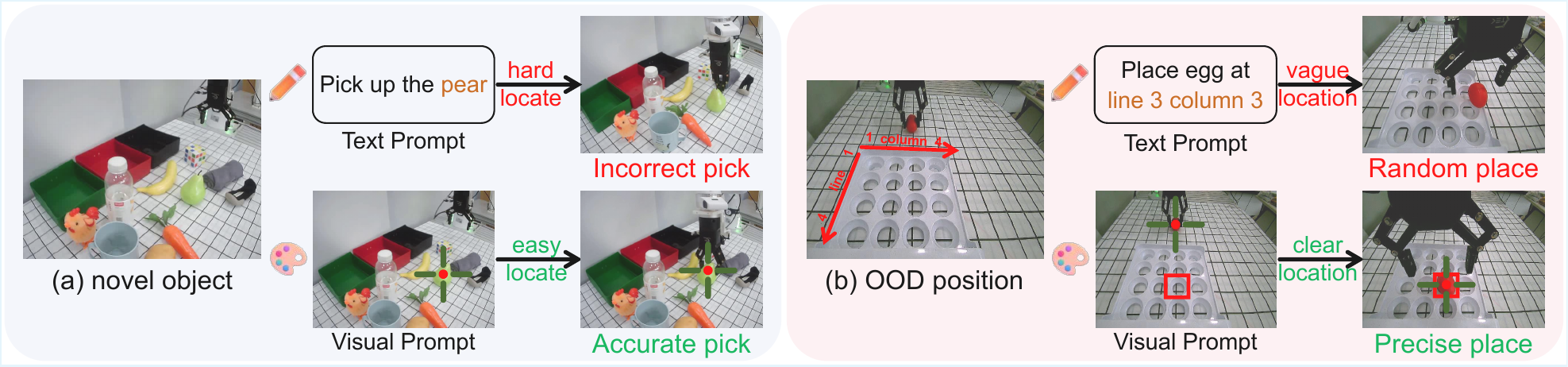}
    \caption{When facing \textbf{(a) novel objects} and \textbf{(b) unseen spatial configurations}, existing VLA models often fail to achieve precise localization (\textcolor{red}{Upper part}), whereas our VP-VLA leverages visual prompts to enhance generalization, ensuring accurate placement for target objects (\textcolor{teal}{Lower part}).}
    \label{fig:teaser_2}
\end{figure}

Despite these successes, existing VLA frameworks often overfit to specific training scene distributions rather than truly grounding instructions in the environment. This is evidenced by recent findings \cite{zhou2025libero, fei2025libero} showing that substituting meaningful language with gibberish barely affects performance. Consequently, these policies often fail when encountering novel object categories or unseen spatial positions, as illustrated in Fig. \ref{fig:teaser_2}. 
To mitigate these issues, several approaches introduce intermediate interfaces, such as goal images \cite{zhao2025cot} or dense geometric supervision \cite{zhang2025dreamvla, zhong2025flowvla}, to provide fine-grained guidance.
However, these methods typically focus on static, single-task scenarios and rely on rigid interface representations. They often fail to account for the dynamic nature of multi-stage tasks, where the required visual focus and affordance should evolve as the task progresses. 
Furthermore, curating dense geometric data for these models is prohibitively expensive, and the quality of predicted affordances remains inconsistent. 
More critically, current VLA systems struggle to effectively integrate high-level reasoning~\cite{kahneman2011thinking} with low-level execution within end-to-end models~\cite{shi2025hi}. 

To address these challenges, we propose VP-VLA, a decoupled dual-system VLA framework. VP-VLA utilizes visual prompts as an explicit, structured interface between high-level reasoning (the ``System 2 Planner'') and low-level execution (the ``System 1 Controller''). Unlike end-to-end models that attempt to implicitly solve instruction interpretation, spatial relation inference, and execution simultaneously, our approach employs a pretrained VLM as a high-level planner. This planner decomposes complex instructions into sub-tasks and identifies relevant target objects and goal locations. These spatial references are then translated into structured visual prompts, including crosshair markers for targets and bounding boxes for placement regions, which are overlaid onto the visual observations for the low-level controller. 
By integrating visual prompts directly in the image space, we transform complex linguistic instructions into precise spatial anchors. To ensure the policy effectively utilizes these cues, we introduce an auxiliary grounding objective. This objective encourages explicit spatial awareness within the VLA controller during training.

We evaluate VP-VLA on diverse simulation benchmarks and real-world scenarios, where it consistently outperforms state-of-the-art methods: 
On Robocasa-GR1-Tabletop benchmark, VP-VLA improves the average success rate by 5\% over the baseline, surpassing competitive models like GR00T-N1.6 \cite{bjorck2025gr00t} without requiring additional large-scale robotic pretraining.
On SimplerEnv benchmark, our method achieves  substantial absolute improvement of +8.3\% over baseline, surpassing prior VLA models including $\pi_{0.5}$ \cite{intelligence2025pi_};
In real-world cluttered scenario, our method consistently yield superior performance on both in-distribution and out-of-distribution evaluations. 
Our contributions are summarized as follows:

\begin{itemize}
\item We propose VP-VLA, a novel framework that decouples high-level reasoning from low-level control through a structured visual prompting interface.
\item We introduce a visual grounding objective during training that enhances the spatial precision and robustness of VLA models.
\item Experiments on Robocasa-GR1-Tabletop, SimplerEnv and real-world scenario demonstrate VP-VLA achieves consistent gains over strong baselines.
\end{itemize}

\section{Related Work}
\label{sec:related}
\textbf{Vision-Language-Action Models.}  Vision-language-action (VLA) models have become a practical paradigm for general-purpose robotic manipulation, translating open-ended semantic instructions into visuomotor policies~\cite{black2024pi0,kim2024openvlaopensourcevisionlanguageactionmodel,xi2024teachingembodiedreinforcementlearning}. 
Leveraging large-scale robot demonstration datasets~\cite{brohan2023rt1roboticstransformerrealworld,o2024open,brohan2023rt2visionlanguageactionmodelstransfer,jones2025sightfinetuninggeneralistrobot,khazatsky2025droidlargescaleinthewildrobot,liu2023liberobenchmarkingknowledgetransfer,walke2023bridgedata}, recent VLAs generalize across diverse tasks and objects by integrating large-scale vision-language models~\cite{bai2025qwen3,he2022galaxygenerativepretrainedmodel,touvron2023llama2openfoundation,openai2024gpt4technicalreport}, multi-modal inputs, and heterogeneous data sources, including real-robot trajectories, human videos, and synthetic simulations. 
However, most methods adopt a monolithic architecture that tightly couples reasoning, spatial grounding, and action generation, hindering task decomposition and intermediate representation~\cite{kim2024openvlaopensourcevisionlanguageactionmodel,black2024pi0,zheng2024tracevla}. Under distribution shifts or personalized scenarios~\cite{lee2025bring}, VLAs remain brittle, particularly for precise instance-level identification or fine-grained spatial reasoning~\cite{geminiroboticsteam2025geminiroboticsbringingai,zawalski2025roboticcontrolembodiedchainofthought,chen2025trainingstrategiesefficientembodied}. These challenges highlight a fundamental gap between high-dimensional sensory observations and sparse, low-dimensional action outputs.

\textbf{Visual Intermediates for Reasoning-Decomposed VLAs.} Prior work has  explored intermediate visual representations to improve spatial reasoning and action grounding in robotic manipulation. Affordance-based methods predict future image frames~\cite{zhao2025cot}, key poses~\cite{Rt-affordance}, trajectories~\cite{Rt-trajectory,li2024coa}, or hierarchical spatiotemporal traces~\cite{zheng2024tracevla,li2025hamster} to inform downstream policies. However, because these intermediate modalities are not natively understood by standard Vision-Language-Action (VLA) models, they typically require task-specific supervision and complex end-to-end training that can compromise the model's inherent reasoning capabilities without guaranteeing executable actions. Seeking a balance between simplicity and information density, recent approaches leverage Vision-Language Models (VLMs) for subtask reasoning and target localization. Training-free pipelines~\cite{tang2025affordgraspincontextaffordancereasoning,ahn2022icanisay} often struggle with low precision due to the VLM's imperfect spatial grounding compared to expert segmentation models. Conversely, methods integrating dense segmentation masks from expert models (e.g., DexGraspVLA~\cite{zhong2026dexgraspvla}, RoboGround~\cite{huang2025roboground}) introduce severe modality mismatches against the original RGB inputs. This necessitates cumbersome architectural modifications—such as training additional vision encoders, designing complex modality fusion layers, or relying on traditional transformer policies like GR-1~\cite{wu2023unleashing}, which significantly increase computational overhead. Other approaches like Point-VLA~\cite{yu2025point} ground objects using static first-frame coordinates, which disconnects the visual prompt from the dynamic manipulation process and relies heavily on explicit image augmentation to prevent severe performance degradation. In contrast, our framework establishes an a native, computationally efficient grounded plan that seamlessly bridges high-level reasoning and low-level execution. By leveraging a pretrained VLM~\cite{bai2025qwen3} for instruction decomposition and SAM3~\cite{carion2025sam3segmentconcepts} for precise target localization, we project interpretable spatial cues directly onto the RGB inputs as lightweight visual overlays. This modality-consistent augmentation provides precise spatial guidance while preserving the VLA's native visual understanding, generalization, and flexibility, effectively eliminating the need for complex modality fusion or additional training overhead.

\section{Method}
\label{sec:method}

\begin{figure}[t]
    \centering
    \includegraphics[width=1.\linewidth]{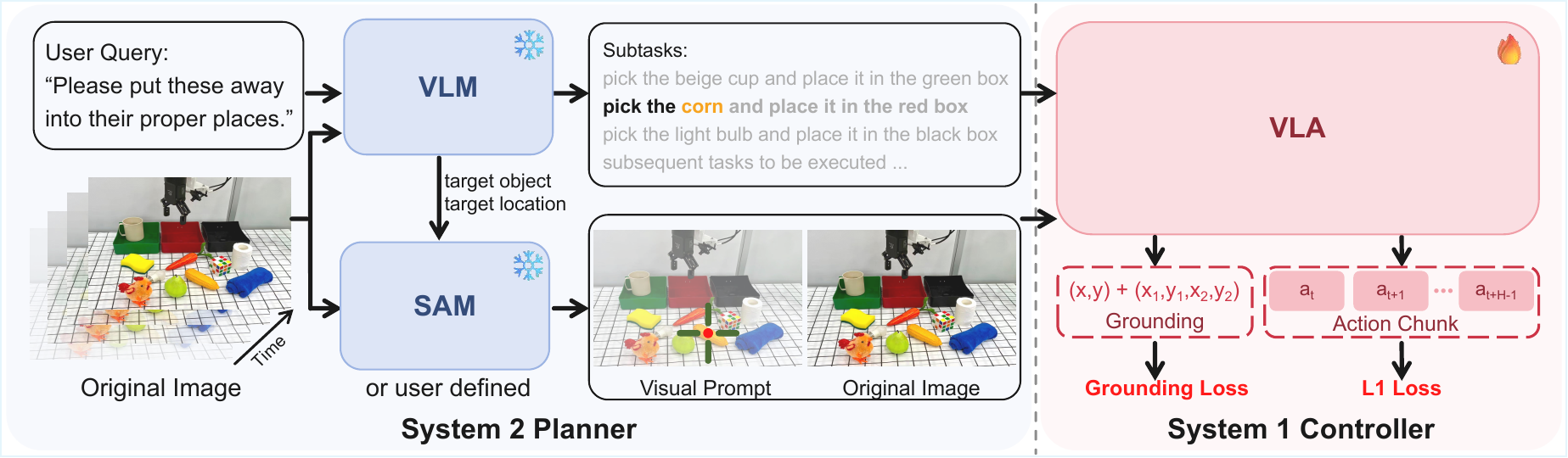}
    \caption{Overview of VP-VLA pipeline. Our VP-VLA leverages a dual-system architecture to bridge high-level reasoning and low-level control. The System 2 planner first decomposes a language instruction into subtasks and generates visual prompts as interaction anchors and spatial constraints. The System 1 controller then utilizes these grounded visual cues to generate precise sensorimotor trajectories for complex, multi-stage manipulation tasks, while a grounding loss aligns the policy with the highlighted regions during training.}
    \label{fig:pipeline}
\end{figure}

We present \textbf{VP-VLA}, a decoupled dual-system framework for robotic manipulation. Following the problem formulation (Sec.~\ref{sec:preliminary}), we propose two core components: (i) the System 2 planner $P_{S2}$ (Sec.~\ref{sec:system2}), an event-driven reasoning module that decomposes tasks into sub tasks and generates visual interface images; and (ii) the System 1 controller $\pi_{\theta}$ (Sec.~\ref{sec:system1}), a high-frequency controller that performs visuomotor tracking conditioned on these visual prompts. 

\subsection{Preliminary}
\label{sec:preliminary}
A standard VLA policy $\pi_{\theta}$ typically maps a language instruction $l$ and a sequence of visual observations $o_t$ to a sequence of action $a_t$ at each time step $t$. The number of visual observation, $o_t = \{ o_t^1, o_t^2, \ldots, o_t^m \}$, which came from a series of overhead or wrist-mounted camera, may vary depending on the embodiment. The sequence of action $a_t = \{ a_t^1, a_t^2, \ldots, a_t^n \}$, called action chunk, will be executed in sequence to compensate the inference delay and keep the execution smooth. $a_t$ will be predicted as follows:
\begin{equation}
a_t = \pi_{\theta}(l, o_t), 
\end{equation}
where $\pi_{\theta}$ is comprised of a pretrained VLM 
and an action decoder typically implemented as an MLP or a diffusion model.

Existing VLA models often suffer from a monolithic bottleneck, where a single network must concurrently manage instruction parsing, spatial reasoning, and motor execution. To address this, we propose VP-VLA, a decoupled dual-system architecture that bridges high-level reasoning and low-level control through an explicit visual interface. 


\subsection{System 2 Planner}
\label{sec:system2}
The high-level System 2 planner, $P_{S2}$, performs deliberative reasoning to obtain the visual interface $I_{vp}^t$. This module operates through two interconnected stages: (i) Event-Driven Task Decomposition, and (ii) Visual Prompt Generation.

\textbf{Event-Driven Task Decomposition.}
Instead of performing computationally expensive high-level reasoning regardless of current progression of tasks~\cite{zhong2025dexgraspvla}, $P_{S2}$ utilizes an event-driven execution loop. We hypothesize that manipulation tasks are composed of discrete semantic phases (e.g., grasp, putting down), and the transitions between these phases are marked by transition events.
We define the transition event $E$ as a change in the robot's physical interaction state $S_t$. Formally, the high-level planner is invoked only when:
\begin{equation}
E_t = \mathds{1} \left( |\phi(S_t) - \phi(S_{t-1})| > \epsilon \right), 
\end{equation}
where $\phi$ is a state-mapping function. In our tabletop manipulation setting, we instantiate $\phi$ as the gripper status. A change in the gripper state (open to closed or vice-versa) serves as a physical proxy for a semantic phase shift, triggering a re-evaluation of the visual prompt to reflect the next sub-goal (e.g., shifting from the target object to the placement destination).

\textbf{Visual Prompt Generation.}
Once an event is triggered, a pretrained VLM planner processes the language instruction $l$ and observation $o_t$, then reason about the subtask that needs to further operate, together with the corresponding target object and target location names from the scene. These names are then passed into a pretrained segmentation model to generate a visual interface image $I_{vp}^t$. This image serves as a spatial bridge, translating abstract language instructions into action affordances. The whole process can be decomposed into semantic reasoning and spatial grounding: In semantic reasoning stage, the planner identifies the current subtask $s_k$ and the associated entities $e \in \{e_{obj}, e_{loc}\}$:\begin{equation}{s_k, e_{obj}, e_{loc}} = \text{VLM}{\text{planner}}(l, o_t, S_t).
\end{equation}
In spatial grounding stage, a segmentation model $\mathcal{G}$ maps these entities to visual prompts $\psi_t$:
\begin{equation}
\psi_t = \mathcal{G}(o_t, e_{obj}, e_{loc}),
\end{equation}
where~$\psi_t$ consists of an interaction anchor~$C \in \mathbb{R}^2$ (denoted as a crosshair) and a spatial constraint~$B \in \mathbb{R}^4$ represented as a bounding box. These visual prompts are then overlaid on the overhead camera observation to obtain~$I_{vp}^t$. Unlike raw images, $I_{vp}^t$~provides explicit geometric priors: for manipulation primitives (e.g., ``pick''), the system generates a crosshair~$C$ at the object's centroid as an anchor for interaction. This reduces the policy's search space from the entire image to a localized region of interaction. For placement primitives, a bounding box~$B$ defines the spatial constraint for target placement. By representing these as explicit visual overlays, we transform the VLA’s task from ``interpreting intent'' to ``visuomotor tracking'' of the provided prompts. After obtaining the visual interface image~$I_{vp}^t$, we feed it together with the original observation~$o_t$ into the System~1 controller~$\pi_{\theta}$.

\subsection{System 1 Controller}
\label{sec:system1}
We extend the standard VLA formulation by introducing the visual prompt image $I_{vp}^t$ at each step, serves as a spatial bridge between the high-level reasoning and grounding and the low-level robot's execution. Our policy is defined as:
\begin{equation}
a_t = \pi_{\theta}(l, o_t, I_{vp}^t).
\end{equation}
The VLA policy $\pi_{\theta}$ consists of a VLM backbone $f_{\omega}$, which processes multimodal inputs into high-level embeddings, and an action decoder $h_{\psi}$, which maps these embeddings to continuous control signals. The policy is thus defined as:
\begin{equation}
a_t = \pi_{\theta}(l, o_t, I_{vp}^t) = h_{\psi} \left( f_{\omega}(l, o_t, I_{vp}^t) \right),
\end{equation}
where $\omega$ and  $\psi$ are the parameters of the VLM and the action decoder, respectively, and $\theta = \{\omega, \psi\}$. 

\textbf{Training Objective.} A key challenge in visual prompting is ensuring the model treats the overlays as semantic anchors rather than extraneous image noise. To address this, we introduce a visual grounding objective that forces the model to internalize the spatial coordinates of the prompts. Our framework can be naturally extended with the auxiliary grounding task. During training, we add an auxillary grounding task on only key frames (first frame and the frame where $E_t = \mathds{1}$). We formulate grounding as a classification task over discretized spatial bins. Following the design of Qwen-3-VL, we divide the image dimensions into $N$ uniform bins, where $N=1000$. For target object crosshair with its center located at $(x, y)$, we query the VLM inside the VLA $\pi_{\theta}$ to predict the 2D location. For target location bounding box, we query the VLM to predict the location $[x_1, y_1, x_2, y_2]$. During training, the VLM $f_{\omega}$ is queried to predict these discretized locations in a structured JSON format. We optimize this using a Cross-Entropy (CE) loss for grounding, which provides a sharper and more structured training signal than traditional MSE. We use L1 loss for action prediction. Critically, the grounding loss is backpropagated only through the VLM parameters $\omega$:
\begin{equation}
\mathcal{L}_{\text{total}} = \mathcal{L}_{\text{action}}(\theta) + \lambda \mathds{1}_{\text{event}} \mathcal{L}_{\text{grounding}}(\omega),
\end{equation}
where~$\lambda$ is the coefficient to balance action prediction and visual prompt grounding. This auxiliary grounding loss ensures that the policy’s internal representations are explicitly aligned with the visual prompts rather than treating them as external noise, leading to more precise and robust manipulation.

\textbf{Data Preparation. } 
For better consistency and efficiency,  we use rule-based approach to first decompose the original task into a subtask list. At key frames, a VLM predicts the current subtask from the list, along with the target object and (if applicable) target location. Using the predicted object and location names, we perform text-conditioned segmentation on all frames to obtain masks and bounding boxes before the next key frame. These annotations are then converted into visual prompts $\psi$: a crosshair placed at the centroid of the target object mask and a bounding box over the target placement region. Each processed episode is stored with per-frame masks, boxes, and VLM subtask records. On Open X-Embodiment (OXE) dataset, we follow the original starVLA~\cite{community2026starvla} experiment setting to discard episodes with empty prompt for both our methods and the baseline.


\section{Experiment}
\label{sec:experiment}
We conduct extensive experiments to validate our method, both in simulation and real-world settings. First, we elaborate the implementation details in Section~\ref{sec:implementation}. Then, we assess performance on the simulation benchmark
in Section~\ref{sec:eval_simulation_robocasa} \&~\ref{sec:eval_simulation_simpler}. Next, we examine real-robot performance on cluttered and under-specified manipulation tasks to study instruction-following and OOD generalization in real-world deployment in Section~\ref{sec:eval_real}.

\subsection{Implementation Details}
\label{sec:implementation}
We use SAM3~\cite{carion2025sam3segmentconcepts} to obtain the visual prompt. We use the default segmentation threshold where detection threshold and mask threshold are 0.5. We keep the visual prompt with highest score for target object and target location respectively. Our codebase is based on starVLA~\cite{community2026starvla} framework, trained on 8 GPUs, and strictly follows the training and evaluation procedure to ensure reproducibility. We adopt QwenOFT architecture, which replace the Prismatic VLM~\cite{karamcheti2024prismatic} in OpenVLA-OFT~\cite{kim2025fine} with Qwen3-VL-4B-Instruct. We use Qwen3-VL-4B-Instruct for System 2 Planner as well. We employ the AdamW optimizer with learning rate as 1e-5 for VLM and 1e-4 for the action model. We set the $\lambda$ to be 0.1 when calculating loss. The visual prompts are all automatically predicted except for the egg carton placement task, where the target object is predicted by the framework and the target location is specified by the user. Baseline performance metrics are sourced from original papers or other peer-reviewed publications. To ensure a fair comparison, the training datasets for these baselines include the data utilized in our own experiments.

\subsection{Experiment on Robocasa Benchmark}
\label{sec:eval_simulation_robocasa}
We applied our pipeline to the Robocasa-GR1-Tabletop benchmark~\cite{nasiriany2024robocasa}, a simulation framework with tabletop kitchen environment, consisting of 24 diverse tasks and in total 24,000 videos. These tasks involves multi-step complex pick and place interactions with varied attributes and geometries. We utilize the Humanoid Robot Tabletop Manipulation subset from the PhysicalAIRobotics-GR00T-X-Embodiment-Sim ~\cite{bjorck2025gr00t} dataset, following ~\cite{community2026starvla, lian2026bayesianvla}. To guarantee reproductibility and statistical significance, we evaluate each task using 50 independent trials and
report the average success rate.

The quantitative results on the RoboCasa Tabletop simulation benchmark are summarized in Table~\ref{tab:robocasa_main_tab}. Taking QwenOFT as the primary baseline, our method achieves a new state-of-the-art average success rate of 53.8\%, outperforming QwenOFT (48.8\%) by a clear margin of +5.0\%. Our approach also surpasses other strong baselines, including Isaac-GR00T N1.5 (48.2\%), Isaac-GR00T N1.6 (47.6\%), QwenGR00T (47.8\%), and QwenPI (43.9\%). Notably, the improvement is particularly evident in the ``PnP * to * Close'' setting, where our method reaches 54.3\%, significantly exceeding QwenOFT (43.7\%) and all other competitors. We observed that for long complex instructions involving multiple steps and nonprehensile grasping, such as ``pick up the wine, place it into the cabinet and close the cabinet'', the VLM reasoner successfully decompose the task into subtask list [\textit{``pick up the wine'', ``place the wine into the cabinet'', ``close the cabinet''}]. In addition, it identifies the target object and the specific affordance required for the final action (the cabinet door). Furthermore, the reasoner accurately detects subtask transitions, ensuring the target object shifts from the ``wine'' to the ``door'' only after the wine has been successfully placed. We also observe consistent gains in several challenging novel generalization splits, such as ``PnP Novel From Placemat To Plate'' (70.0\% vs. 52.0\% for QwenOFT) and ``PnP Novel From Tray To Plate'' (66.0\% vs. 56.0\%), where the evaluation includes random initialized position, novel appearance, and distracting object and container. Our method not only improves overall task success rate but also enhances generalization for varing background, object attribute and position. 

\begin{table*}[!t]
    \centering
    \small
    \renewcommand{\arraystretch}{1.4} 
    \setlength{\tabcolsep}{1.6pt} 

    \caption{
      Results of evaluating the VLA models with the GR1 robot in the RoboCasa Tabletop simulation environment. We highlight the best result in ~\textbf{bold} and the second-best results with \underline{underline}.
    }
     \begin{adjustbox}{width=0.9\linewidth}
    \begin{tabular}{l c c c c c c c}
        \toprule
        \rowcolor{white} 
        {Task} & 
        {\scriptsize \makecell{\textbf{Isaac-GR00T}\\\textbf{N1.5}}} & 
        {\scriptsize \makecell{\textbf{Isaac-GR00T}\\\textbf{N1.6}}} & 
        {\scriptsize \makecell{\textbf{QwenGR00T}\\\textbf{+Qwen3VL}}} & 
        {\scriptsize \makecell{\textbf{QwenPI}\\\textbf{+Qwen3VL}}} & 
        {\scriptsize \makecell{\textbf{QwenOFT}\\\textbf{+Qwen3VL}}} & 
        {\scriptsize \makecell{\textbf{QwenFAST}\\\textbf{+Qwen3VL}}} &
        {\scriptsize \makecell{\textbf{Ours}\\\textbf{+Qwen3VL}}}\\
        \midrule
        PnP * to * Close (Avg)                       & 45.3 & 24.2 & 50.3 & 42.3 & 43.7 & 35.0 & 54.3\\
        PnP Novel From Cuttingboard To * (Avg)       & 46.4 & 56.9 & 52.8 & 46.0 & 50.4 & 50.4 & 60.8\\
        PnP Novel From Placemat To * (Avg)           & 45.5 & 51.9 & 38.0 & 43.5 & 41.5 & 33.5 & 54.5\\
        PnP Novel From Tray To * (Avg)               & 48.8 & 55.1 & 39.2 & 44.0 & 49.2 & 32.0 & 46.0\\
        PnP Novel From Plate To * (Avg)              & 56.5 & 57.6 & 58.5 & 44.0 & 61.0 & 45.0 &  53.5\\
        \rowcolor{gray!30} 
        \textbf{Average}                                                      & 48.2 & 47.6 & 47.8 & 43.9 & \underline{48.8} & 39.0 & \textbf{53.8}\\
        \bottomrule
    \end{tabular}
        \end{adjustbox}
    \label{tab:robocasa_main_tab}
\end{table*}

\subsection{Experiment on SimplerEnv Benchmark}
\label{sec:eval_simulation_simpler}
We utilize two large-scale subsets from the Open X-Embodiment (OXE) dataset: BridgeDataV2 ~\cite{walke2023bridgedata} and Fractal ~\cite{brohan2023rt1roboticstransformerrealworld}. The model is fine-tuned for 70k steps on 8 GPUs (batch size 32 per device). This benchmark includes four manipulation tasks: ``Put spoon on towel'', ``Put carrot on plate'', ``Stack green cube on yellow cube'', and ``Put eggplant in yellow basket''. We evaluate the policies using the official evaluation scripts provided by the SimplerEnv repository~\cite{li2024evaluating}.
\definecolor{navyblue}{HTML}{0071BC}

The quantitative results on the SimplerEnv simulation benchmark are summarized in Table ~\ref{tab:simplerenv}. Using QwenOFT as the primary baseline (50.0\% average), our method achieves a new state-of-the-art performance of 58.3\%, yielding a substantial improvement of +8.3\%. Compared with other strong competitors, our approach also surpasses $\pi_{0.5}$ (57.1\%) and Isaac-GR00T-N1.6-Bridge (57.1\%), and outperforms prior VLA systems such as CogACT (51.3\%) and VideoVLA (53.1\%). At the task level, we observe notable improvements over QwenOFT in tasks requiring precise object identification, manipulation and target location grounding, including ``Put Spoon on Towel'' (66.7\% vs. 58.3\%) and a substantial gain in ``Put Eggplant in Yellow Basket'' (95.8\% vs. 70.8\%). These findings suggest that our approach more effectively leverages language-conditioned signals to guide action selection, establishing a new performance ceiling on this benchmark.

\begin{table*}[!t]
  \centering
  \caption{
     Results of evaluating the VLA models with the WidowX robot in the SimplerEnv simulation environment. We highlight the best results in \textbf{bold} and the second-best results with \underline{underline}.
    }
  \begin{adjustbox}{width=0.9\linewidth}
  \rowcolors{24}{white}{gray!15}
  \begin{tabular}{l c c c c c}
    \toprule
    \textbf{Method}
     & \makecell[c]{\textbf{Put Spoon} \\ \textbf{on Towel}} 
     & \makecell[c]{\textbf{Put Carrot} \\ \textbf{on Plate}} 
     & \makecell[c]{\textbf{Stack Green Block} \\ \textbf{on Yellow Block}} 
     & \makecell[c]{\textbf{Put Eggplant} \\ \textbf{in Yellow Basket}} 
     & \textbf{Average} \\
    \midrule
    RT-1-X~\cite{o2024open}         &  0.0  & 4.2   & 0.0   & 0.0   & 1.1 \\
    Octo-Base~\cite{team2024octo}       & 15.8  & 12.5  & 0.0   & 41.7  & 17.5 \\
    Octo-Small~\cite{team2024octo}      & 41.7  & 8.2   & 0.0   & 56.7  & 26.7 \\
    OpenVLA-OFT~\cite{kim2025fine}     & 34.2  & 30.0  & 30.0  & 72.5  & 41.8 \\
    RoboVLM~\cite{liu2025towards}         & 50.0  & 37.5  & 0.0   & 83.3  & 42.7 \\ 
    Magma~\cite{yang2025magma}                & 37.5  & 29.2  & 20.8  & 91.7  & 44.8 \\  
    CogACT~\cite{li2024cogact}           & 71.7 &  50.8  & 15.0 & 67.5 & 51.3 \\
    SpatialVLA~\cite{qu2025spatialvla}      & 20.8  & 20.8  & 25.0  & 70.8  & 34.4 \\
    TraceVLA~\cite{zheng2024tracevla}        & 12.5  & 16.6  & 16.6  & 65.0  & 27.7 \\
    VideoVLA~\cite{shen2025videovla}        & 75.0 & 20.8   & 45.8 & 70.8 & 53.1 \\
    \midrule
    $\pi_0$~\cite{black2024pi0}         & 29.2 & 62.5 & 29.2 & 91.6 & 53.1 \\
    $\pi_{0.5}$~\cite{intelligence2025pi_} & 49.3 & 64.7 & 44.7 & 69.7 & \underline{57.1} \\
    Isaac-GR00T-N1.6-Bridge~\cite{bjorck2025gr00t}   & 64.5 & 65.5 & 5.5 & 93.0 & \underline{57.1} \\
    \midrule
    QwenOFT + Qwen3VL & 58.3 & 50.0 & 20.8 & 70.8 & 50.0\\
    Ours + Qwen3VL  & 66.7 & 50.0 & 20.8 & 95.8 & \textbf{58.3}\\
    \bottomrule
  \end{tabular}
  \end{adjustbox}
  \label{tab:simplerenv}
\end{table*}

\subsection{Experiment on Real-world Scenario}
\label{sec:eval_real}
We comprehensively evaluate VP-VLA across multiple real-world manipulation tasks to validate its core capabilities in several dimensions. Specifically, we focus on: (i) the reasoning and grounding ability within cluttered scenes reflected on overall success rates; (ii) the robustness and generalization ability in out-of-distribution (OOD) settings; and (iii) the effectiveness of visual prompting against textual prompting in scenarios requiring complex spatial reasoning.

\begin{figure}[t]
    \centering
    \includegraphics[width=.9\linewidth]{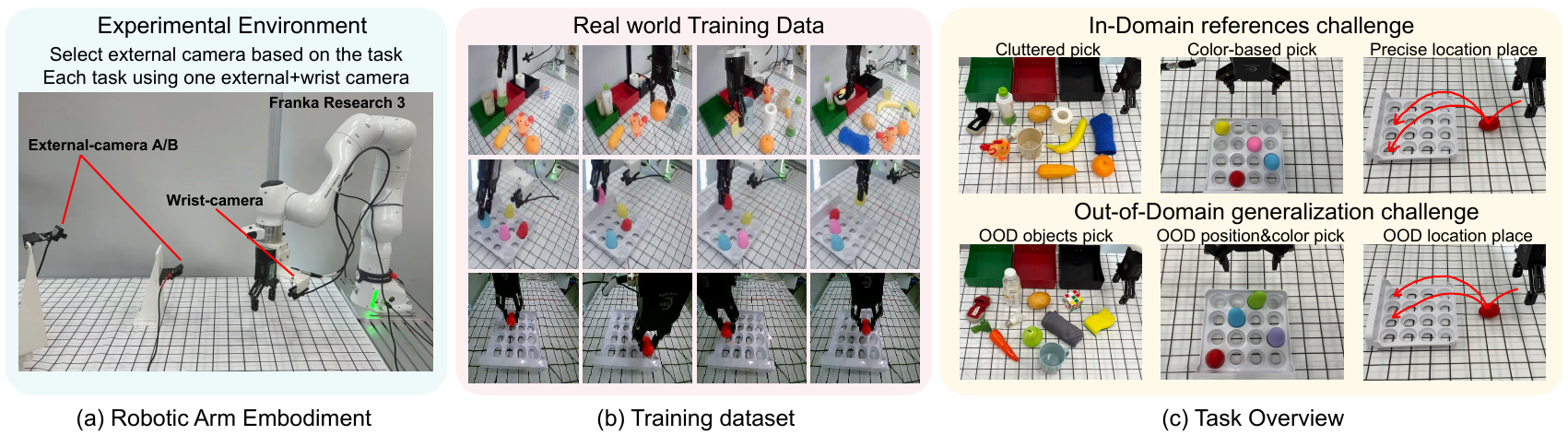}
    \caption{Overview of our real-world task and robot setting. (a) Robot setting for the experiments. We use an external camera A for categorization task and pick colored egg task, while using the external camera B for egg carton placement task. (b) We collect real-world robot demonstration for three task suites. (c) We present examples of each task with the OOD setting illustration.}
    \label{fig:real-setting}
\end{figure}

For each of the following section, we introduce our experimental setup and evaluation results. We use a stationary, table-mounted Franka Research 3 7-DoF robot arm. Environment observation includes two RGB images: one from a fixed third-person perspective and the other from a first-person camera mounted on the end-effector. The setup is shown in ~\ref{fig:real-setting}. Both images are resized to 224 × 224 before feeding into the model. We fine-tune both the baseline model and our method using 8 GPUs, with total batch size of 256 and action chunk size of 16. 

\textbf{Robotic Waste-Sorting Categorization.} 
To evaluate visual grounding and generalization in cluttered environments, we design a robotic pick-and-place task inspired by daily waste-sorting. Models must follow instructions to categorize randomly posed objects into specific containers, which requires both visual grounding ability and precise action prediction. We collect 50 trajectories per training object and evaluate both our method and the QwenOFT baseline at 10K training steps. We evaluate performance across In-Domain (ID) setting and out-of-distribution setting (OOD) with novel object.

\begin{figure}[t]
    \centering
    \includegraphics[width=.8\linewidth]{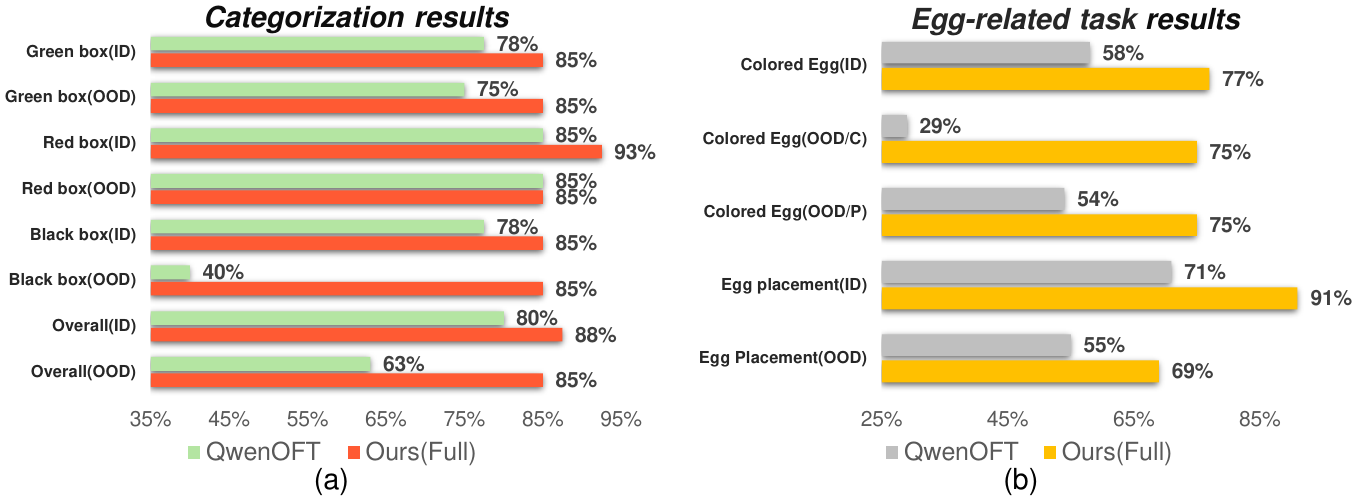}
    \caption{Results of real-world robot manipulation. (a) Categorization result; (b) Pick colored egg result and egg carton placement result.
    }
    \label{fig:real_world_result}
\end{figure}

As shown in Fig. \ref{fig:real_world_result}, our method achieves an 87.5\% ID success rate and 85\% OOD success rate, significantly outperforming QwenOFT (80\% ID, 63.3\% OOD). Notably, our method exhibits a minimal generalization gap (2.5\%), whereas QwenOFT suffers a 16.7\% performance drop. This suggests that while the baseline overfits to the training distribution, our approach maintains robust object-level grounding across categories in cluttered scenes.


\textbf{Object Reference by Attribute.}
To evaluate fine-grained attribute grounding, we design a color-based reference task: ``pick up the \textit{<color>} egg.'' Unlike category-level reasoning in the previous experiment, this task focuses on attribute-level grounding, requiring the model to bind specific linguistic color tokens to visual instances under spatial variation. Using a $4\times4$ grid, we collect 200 demonstrations (50 per color for blue, pink, red, and yellow). We evaluate performance across three settings: In-Domain (ID), OOD Color (novel purple and green eggs), and OOD Position (eggs placed in 12 unseen grid locations).

Fig.~\ref{fig:real_world_result} summarizes the performance across all settings. Our method significantly outperforms the baseline, particularly under distribution shifts. For in-domain scenario, our model achieves $77.1\%$ accuracy, outperforming the baseline by a large margin of $18.8\%$. For OOD Color scenario, our method maintains $75.0\%$ success on novel colors, while the baseline drops to $29.2\%$. The minimal gap between our ID and OOD color performance ($2.1\%$) suggests the learned representation captures attribute semantics rather than memorizing instances. In contrast, the baseline exhibits severe overfitting to seen color distributions. Finally, for OOD position with unseen grid locations, our model reaches $75.0\%$ compared to $54.2\%$ for the baseline, demonstrating spatial generalization beyond training data, while the baseline shows weaker position invariance.

\textbf{Location reference with spatial grounding.}
In extreme scenario where it is hard to specify the exact location, we present an option to let the user provide the visual prompt. To evaluate whether our proposed method can still follow user’s intent in such scenario, we design an egg carton placement task. The robot is tested to pick up an egg and place it at a language-specified coordinate (e.g., ``line 2, column 4'') on a 4$\times$4 grid. We manually drawn visual prompt (bounding box) at the target location for our method to resolve potential linguistic ambiguity, whereas the baseline must ground the instruction purely from text. We evaluate on In-Domain (ID) and Out-of-Distribution (OOD) coordinates using a partial credit system: 1.0 for the target cell, 0.5 for adjacent, and 0.25 for diagonal cells.

As shown in Fig.~\ref{fig:real_world_result}, our method consistently outperforms QwenOFT across all settings: For In-Domain setting, our model achieves 91.25\% accuracy (36.5/40), a more than 20\% improvement over QwenOFT (70.63\%). This reflects stable spatial grounding when coordinates are seen during training. For out-of-distribution scenario with novel row–column combinations, our method reaches 68.75\% , compared to 55\% for the baseline. While OOD compositional generalization is challenging for both, our model maintains a significant performance lead, suggesting superior geometric grounding over memorization of frequently seen coordinates.

\subsection{Ablation Study}
\label{sec:ablation}
We conduct ablation experiments on the RoboCasa Tabletop simulation to analyze the contribution of each design component. Table~\ref{tab:robocasa_ablation_styled} reports the results.

\textbf{Effect of the Grounding Objective.}
Removing the grounding loss (``w/o grounding'') reduces the overall success rate from 53.8\% to 49.4\%, demonstrating that explicitly aligning policy representations with the prompted regions is crucial. Without this constraint, the model may perceive the visual prompts but fail to consistently associate them with action generation. We further evaluate grounding applied to every frame (``w/ all frame grounding''). This variant achieves 49.5\%, which is lower than our key-frame grounding strategy. Applying grounding supervision densely across all frames may introduce redundant or noisy constraints, leading to unstable training and suboptimal optimization. This result suggests that selective grounding at key decision frames provides a better balance between supervision strength and training stability.

\textbf{Effect of Visual Prompt Design.}
Changing the target object prompt from a crosshair to a point (``w/ point'') degrades performance to 47.3\% on average. A single point provides weaker spatial extent information compared to a structured crosshair, making it harder for the policy to infer object location but rather treating it as visual perturbation. This indicates that prompt geometry influences how effectively spatial cues are interpreted. We also test directly overlaying prompts on the primary RGB image (``w/ direct overlay''), which yields 50.8\%. Separating the visual prompt avoids excessive interference with raw visual features and prevent overlapping.

\textbf{Overall Discussion.}
Our full method ranked the best among other variants. The results demonstrate that (i) grounding supervision is necessary for reliable prompt utilization, (ii) supervision should be applied selectively rather than densely, and (iii) prompt representation design significantly affects perception. Together, these findings validate the current design.

\begin{table}[!t]
    \centering
    \scriptsize
    \renewcommand{\arraystretch}{1.4}
    \caption{Ablation results of evaluating the VLA models with the GR1 robot in the RoboCasa Tabletop simulation environment}
    \begin{tabularx}{\linewidth}{>{\raggedright\arraybackslash}X ccccc}
        \toprule
        \textbf{Task} & 
        {\scriptsize \makecell{\textbf{w/o}\\\textbf{grounding}}} & 
        {\scriptsize \makecell{\textbf{w/ all frame}\\\textbf{grounding}}} & 
        {\scriptsize \makecell{\textbf{w/}\\\textbf{point}}} & 
        {\scriptsize \makecell{\textbf{w/ direct}\\\textbf{overlay}}} & 
        {\scriptsize \makecell{\textbf{Ours}\\\textbf{(Full)}}} \\
        \midrule
        
        PnP * to * Close (Avg) & 49.7 & 46.3 & 38.7 & 47.7 & 54.3 \\
        PnP Novel From Cuttingboard To * (Avg) & 54.4 & 53.6 & 48.8 & 54.0 & 60.8 \\
        PnP Novel From Placemat To * (Avg) & 46.0 & 45.0 & 46.0 & 46.5 & 54.5 \\
        PnP Novel From Tray To * (Avg) & 43.2 & 46.0 & 49.6 & 50.0 & 46.0 \\
        PnP Novel From Plate To * (Avg) & 54.0 & 58.0 & 56.5 & 56.5 & 53.5 \\
        
        \rowcolor{gray!30}
        \textbf{Average} & 49.4 & 49.5 & 47.3 & 50.8 & 53.8 \\
        \bottomrule
    \end{tabularx}
    \label{tab:robocasa_ablation_styled}
\end{table}


\section{Conclusion}
\label{sec:conclusion}
This paper presents VP-VLA, a novel dual-system vision-language-action framework. By decoupling high-level reasoning from low-level execution, our approach leverages a ``System 2 Planner'' to translate complex linguistic instructions into explicit and structured visual prompts. These spatial anchors, combined with an auxiliary visual grounding objective during training, effectively guide the ``System 1 Controller'' to perform precise manipulation tasks. 
Extensive evaluations across simulated benchmarks, including Robocasa-GR1-Tabletop and SimplerEnv, as well as real-world cluttered scenarios, demonstrate the superior performance and generalization of VP-VLA.


\clearpage
\bibliographystyle{unsrtnat}
\bibliography{references}


\appendix
\newpage
\section{Limitations}
Our interface currently utilizes 2D bounding boxes and crosshair coordinates as spatial anchors. While these primitives are sufficient for a wide range of robotic tasks, more complex manipulations, such as threading a needle or handling deformable objects, might benefit from richer visual prompts, such as 3D Gaussian splats or contact-point heatmaps. Exploring the optimal balance between prompt simplicity and information density is a promising direction for the community.

\section{Compute Resources}
All experiments were conducted on a server equipped with eight NVIDIA H200 GPUs, each with 140 GB of memory. Real world inference is conducted on a single NVIDIA 4090 with 24 GB memory. Training on the full dataset took approximately 46 hours.

\section{Inference Latency}
We further conduct experiments on system latency. We tested with Qwen3-VL-8B-Instruct as the VLM planner with OpenRouter service and locally host SAM3 server. We achieves a mean latency of 0.78s per reasoning call, while the action policy infers each action chunk in 0.108s on average. Meanwhile, SAM3 segmentation runs at 0.36s on average, remaining substantially faster than semantic VLM planning and compatible with asynchronous prompt updates.

\section{Additional Demonstration to Tool-Use Manipulation Tasks}
Tool-use tasks are inherently long-horizon and stage-dependent. To support the claim that our method can be applied beyound pick-and-place tasks, we conduct a study on tool-use tasks. We design a scoop bean task that requires the robot arm to first pick up the spoon, scoop on one bowl and then pour it on another bowl. Given each episode, we first detect subtask boundaries from end-effector motion. Unlike gripper-state heuristics, we use z-axis rise events as transition cues, which is crucial in tasks where the gripper remains mostly closed (e.g., spoon manipulation). We collected in total 106 episodes for training and trained both the baseline model and our methods for 50k steps with batch size 32.

\begin{figure}[t]
    \centering
    \includegraphics[width=.8\linewidth]{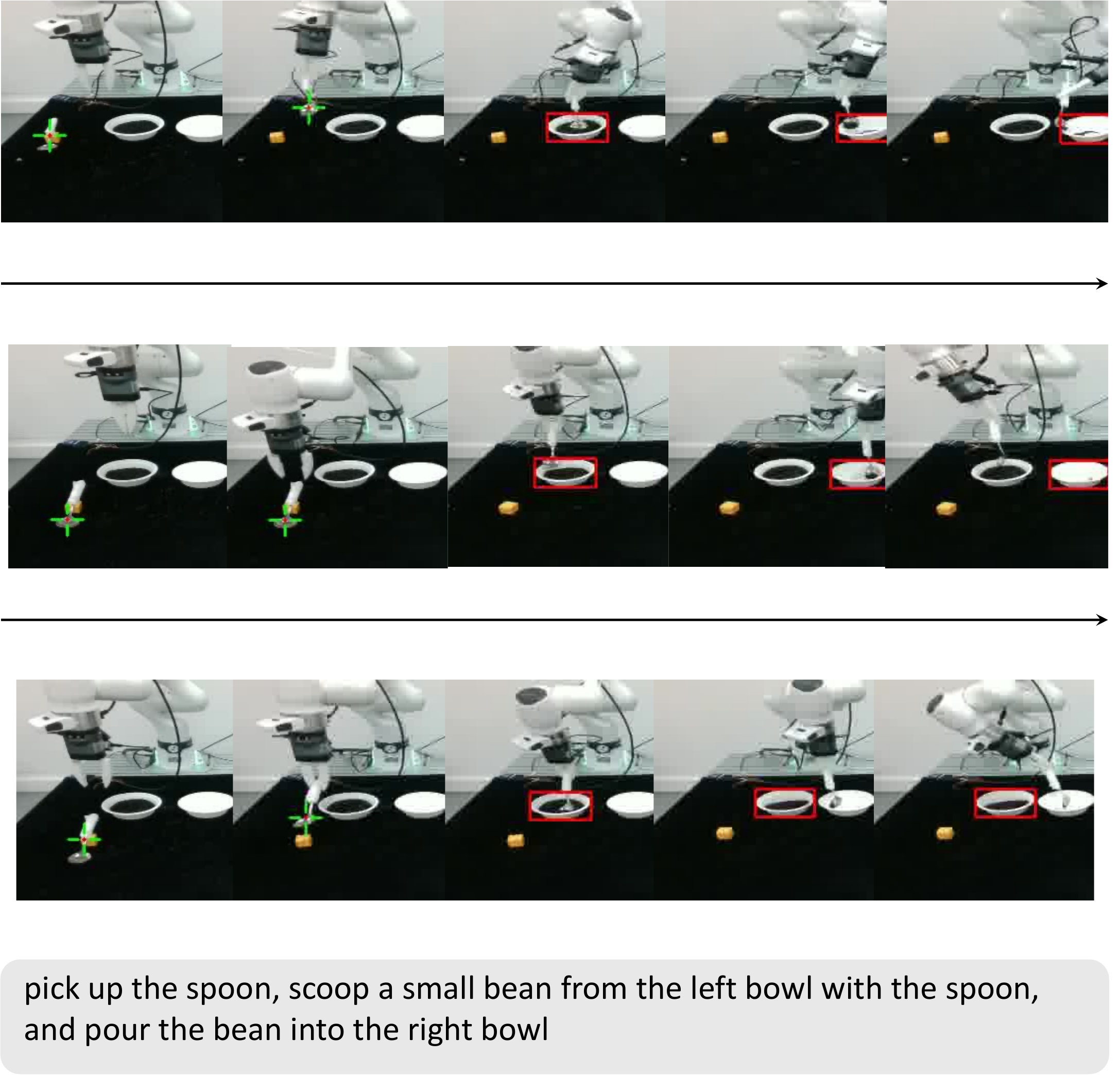}
    \caption{Inference visualization on real-world tasks.}
    \label{fig:appendix_scoop}
\end{figure}

Same as previous setting, at the first frame and each detected boundary, a VLM receives a reference frame and the current frame, then predicts whether to continue or proceed to the next subtask, together with target object and (if needed) target location. These semantic targets are subsequently grounded by a segmentation model (SAM), producing frame-level masks/boxes across the episode. The resulting annotations are saved as compact per-episode files and injected into training via a visual-prompt dataset: the policy conditions on the current observation augmented with object/location prompts. This creates a closed loop from high-level task progress reasoning to pixel-level action-relevant grounding. During experiment, we noticed the baseline struggles for moving towards the right bowl for pouring, while ours explicitly clue the model to proceed with visual prompts. Result are shown in Table~\ref{tab:scoop_success_rate}. An illustration on the visual prompt are shown in Fig.~\ref{fig:appendix_scoop}.

\begin{table}[ht]
\centering
\caption{Comparison of Scoop Beans Performance}
\label{tab:scoop_success_rate}
\begin{tabular}{lc}
\toprule
\textbf{Models} & \textbf{Success Rate (\%)} \\
\midrule
QwenOFT         & 30 \\
Ours            & \textbf{50} \\
\bottomrule
\end{tabular}
\end{table}

\section{Extended Analyze on Experiments}
\textbf{Full Per Task Results on RoboCasa.} Table~\ref{tab:app_robocasa_main_tab} presents the per-task success rates for the RoboCasa
benchmark. This detailed view expands upon the main study, highlighting how our method performs across diverse tabletop manipulation tasks.

\textbf{Robotic Waste-Sorting Categorization.}  In Table~\ref{tab:app_categorization_results}, we further analyze the performance for each boxes: (i) In the Green Box (Recyclable), QwenOFT struggled with a red shoe (7/10), suggesting a reliance on color heuristics. Our model remained robust (9/10), demonstrating category grounding independent of surface appearance. (ii) For the Red Box (Kitchen Waste), our model maintained 85\% accuracy on novel OOD objects (pear, carrot) compared to 75\% for the baseline, reflecting superior semantic abstraction. (iii) The Black Box (Other Waste) highlighted the largest gap. On scrambled Rubik’s cubes, which share semantic identity with the training set but differ visually, QwenOFT’s performance collapsed (3/10) due to imprecise grasping and pattern overfitting. Our method achieved 9/10 by strictly following visual prompts and grounding the object to the correct placement location, even for challenging OOD items like sponge (8/10 vs. 5/10), where the baseline appears not knowing which box to place the object.

\textbf{Object Reference by Attribute.} In Table~\ref{tab:app_colored_egg_results}, we further analyze the performance for each configuration: (i) ID Colors: our method consistently surpasses baseline across all training colors. Failures in baseline often involve imprecise grasp positioning or failing to move above the target-colored egg, indicating weaker visuomotor alignment for attribute-conditioned manipulations. (ii) OOD Color: The baseline frequently confuses novel colors with visually similar training colors or defaults to the egg closest to the gripper while ignoring the color condition. In contrast, our model consistently moves accurately to the target, achieving $9/12$ for both colors. This suggests better disentanglement between linguistic attributes and spatial proximity bias. (iii) OOD Position: For eggs placed at unseen positions, the baseline shows a bias toward previously demonstrated columns, indicating spatial memorization. Our model achieves $9/12$ for both colored eggs, demonstrating spatial generalization and more stable grasping under positional shift.

\textbf{Location reference with spatial grounding.} In Table~\ref{tab:app_egg_grid_results}, we further analyze the performance for each configuration: (i) In-Domain Positions: Our model achieves near-perfect scores on most ID coordinates, indicating accurate row–column parsing. Conversely, QwenOFT shows high variance; for instance, it drops to 1/5 at L3C3 (vs. our 4.5/5), struggling to resolve spatial references when vertical and horizontal axes must be composed jointly. (ii) Out-of-Distribution Positions: OOD tasks require the model to follow target masks at coordinates never jointly observed during training. Our method maintains strong performance at L3C2 (5/5) and L4C3 (4/5), successfully extending the visual-prompt-following mechanism to new locations. QwenOFT remains inconsistent (1.75/5 to 3.5/5), often failing to generalize the underlying grid geometry to novel index combinations.

\section{Extended Ablation Results}
In this section, we provide a comprehensive breakdown of our experimental evaluations to further validate the design choices of VP-VLA.

\textbf{Full Ablation Results on RoboCasa.} Table~\ref{tab:robocasa_ablation_detail} presents the per-task success rates for the RoboCasa benchmark. This detailed view expands upon the main ablation study, highlighting how our design choice performs across diverse tabletop manipulation tasks.

\begin{table*}[!t]
    \centering
    \small
    \renewcommand{\arraystretch}{1.4} 
    \setlength{\tabcolsep}{1.6pt} 

    \caption{
      Per-task result for the ablation on Robocasa.
    }
     \begin{adjustbox}{width=\linewidth}
    \begin{tabular}{l c c c c c}
        \toprule
        \rowcolor{white} 
        {Task} & 
{\scriptsize \makecell{\textbf{w/o}\\\textbf{grounding}}} & 
        {\scriptsize \makecell{\textbf{w/ all frame}\\\textbf{grounding}}} & 
        {\scriptsize \makecell{\textbf{w/}\\\textbf{point}}} & 
        {\scriptsize \makecell{\textbf{w/ direct}\\\textbf{overlay}}} & 
        {\scriptsize \makecell{\textbf{Ours}\\\textbf{(Full)}}} \\
        \midrule
        PnP Bottle To Cabinet Close &                                 52.0 & 34.0 & 40.0 & 46.0 & 54.0 \\
        PnP Can To Drawer Close &                                     70.0 & 66.0 & 42.0 & 78.0 & 72.0\\
        PnP Cup To Drawer Close  &                                    48.0 & 54.0 & 32.0 & 48.0 & 44.0\\
        PnP Milk To Microwave Close &                                 52.0 & 40.0 & 48.0 & 48.0 & 74.0\\
        PnP Potato To Microwave Close &                               42.0 & 46.0 & 30.0 & 26.0 & 34.0\\
        PnP Wine To Cabinet Close &                                   34.0 & 38.0 & 40.0 & 40.0 & 48.0\\
        \midrule
        \rowcolor{gray!20}\textbf{PnP * to * Close (Avg)} &          49.7 & 46.3 & 38.7 & 47.7 & 54.3   \\
        \midrule
        PnP Novel From Cuttingboard To Basket &                      50.0 & 60.0 & 58.0 & 52.0 & 66.0\\
        PnP Novel From Cuttingboard To Cardboardbox  &               58.0 & 50.0 & 44.0 & 58.0 & 54.0\\
        PnP Novel From Cuttingboard To Pan &                         64.0 & 62.0 & 52.0 & 60.0 & 74.0\\
        PnP Novel From Cuttingboard To Pot &                         64.0 & 56.0 & 48.0 & 54.0 & 54.0\\
        PnP Novel From Cuttingboard To Tieredbasket   &              36.0 & 40.0 & 42.0 & 46.0 & 56.0\\
        \midrule
        \rowcolor{gray!20}\textbf{PnP Novel From Cuttingboard To * (Avg)} & 
           54.4 & 53.6 & 48.8 & 54.0 & 60.8\\
        \midrule
        PnP Novel From Placemat To Basket  &                         56.0 & 36.0 & 44.0 & 48.0 & 48.0 \\
        PnP Novel From Placemat To Bowl  &                           48.0 & 46.0 & 46.0 & 60.0 & 74.0\\
        PnP Novel From Placemat To Plate  &                          56.0 & 66.0 & 70.0 & 60.0 & 70.0\\
        PnP Novel From Placemat To Tieredshelf  &                    24.0 & 32.0 & 24.0 & 18.0 & 26.0\\
        \midrule
        \rowcolor{gray!20}\textbf{PnP Novel From Placemat To * (Avg)} &
           46.0 & 45.0 & 46.0 & 46.5 & 54.5\\
        \midrule
        PnP Novel From Tray To Cardboardbox &                        40.0 & 46.0 & 58.0 & 46.0 & 44.0\\
        PnP Novel From Tray To Plate  &                              52.0 & 56.0 & 56.0 & 66.0 & 66.0\\
        PnP Novel From Tray To Pot  &                                56.0 & 56.0 & 62.0 & 58.0 & 38.0\\
        PnP Novel From Tray To Tieredbasket &                        46.0 & 46.0 & 42.0 & 58.0 & 58.0\\
        PnP Novel From Tray To Tieredshelf  &                        22.0 & 26.0 & 30.0 & 22.0 & 24.0\\
        \midrule
        \rowcolor{gray!20}\textbf{PnP Novel From Tray To * (Avg)}& 
           43.2 & 46.0 & 49.6 & 50.0 & 46.0\\
        \midrule
        PnP Novel From Plate To Bowl &                               58.0 & 70.0 & 60.0 & 54.0 & 52.0\\
        PnP Novel From Plate To Cardboardbox   &                     38.0 & 38.0 & 46.0 & 38.0 & 44.0 \\
        PnP Novel From Plate To Pan  &                               54.0 & 50.0 & 58.0 & 60.0 & 56.0\\
        PnP Novel From Plate To Plate &                              66.0 & 74.0 & 62.0 & 74.0 & 62.0\\
        \midrule
        \rowcolor{gray!20}\textbf{PnP Novel From Plate To * (Avg)}&
           54.0 & 58.0 & 56.5 & 56.5 & 53.5\\
        \midrule
        \rowcolor{gray!30} 
        \textbf{Average} &                                           49.4 & 49.5 & 47.3 & 50.8 & 53.8      \\
        \bottomrule
    \end{tabular}
        \end{adjustbox}
    \label{tab:robocasa_ablation_detail}
\end{table*}

\textbf{Impact of Reasoning Decomposition.} We conducted additional ablation studies in the SimplerEnv environment to assess the necessity of temporal decomposition. Our results in Table~\ref{tab:simplerenv_ablation} indicate that models lacking decomposition, where both the interaction anchor (crosshair) and spatial constraint (bounding box) are rendered simultaneously, suffer a degradation in success rate, especially the ``Put Eggplant
in Yellow Basket'' task. We hypothesize that concurrent prompts introduce visual noise that confuses the policy’s attention. Furthermore, prompting without decomposition would fail to generalize to complex, multi-step sequences, such as the ``PnP * to * Close'' task suite, where the system must distinguish between sequential objectives.

\begin{table*}[!t]
  \centering
  \caption{
     Ablation result on applying task decomposition on SimplerEnv.
    }
  \begin{adjustbox}{width=\linewidth}
  \rowcolors{24}{white}{gray!15}
  \begin{tabular}{l c c c c c}
    \toprule
    \textbf{Method}
     & \makecell[c]{\textbf{Put Spoon} \\ \textbf{on Towel}} 
     & \makecell[c]{\textbf{Put Carrot} \\ \textbf{on Plate}} 
     & \makecell[c]{\textbf{Stack Green Block} \\ \textbf{on Yellow Block}} 
     & \makecell[c]{\textbf{Put Eggplant} \\ \textbf{in Yellow Basket}} 
     & \textbf{Average} \\
    \midrule
    Ours w/o decomposition & 70.8 & 62.5 & 16.7 & 79.2 & 57.3 \\
    Ours + Qwen3VL  & 66.7 & 50.0 & 20.8 & 95.8 & \textbf{58.3}\\
    \bottomrule
  \end{tabular}
  \end{adjustbox}
  \label{tab:simplerenv_ablation}
\end{table*}

\section{System 2 Planner Prompt Details}
The prompt templates for the System 2 planner are provided in Table~\ref{tab:task_decomposition_prompt} and Table~\ref{tab:subtask_detection_prompt}. The planner first decomposes a high-level language instruction into an ordered list of atomic subtasks. During execution, the planner is selectively invoked by transition events to re-evaluate the current stage and determine whether to proceed to the subsequent subtask. Upon triggering, the planner identifies the corresponding target object and destination names within the scene. Empirically, we find that incorporating the current gripper state into the planner’s input context enhances its decision-making accuracy regarding task completion.

\begin{table*}[t]
\centering
\small
\setlength{\tabcolsep}{6pt}
\renewcommand{\arraystretch}{1.5}
\caption{Prompt structure used for task decomposition. The VLM planner decomposes a high-level task description into atomic robotic manipulation subtasks.}
\label{tab:task_decomposition_prompt}

\begin{tabularx}{\textwidth}{lX}
\toprule
\textbf{Section} & \textbf{Prompt Content} \\
\midrule

Task Description & \texttt{\{task\_description\}} \\

Instruction & Decompose this robotic manipulation task into sequential atomic subtasks. \\

Subtask Types & 
\mbox{}\begin{itemize}[nosep, leftmargin=*, after=\vspace{-\baselineskip}]
    \item Pick action (e.g., ``pick up the cup'')
    \item Place action (e.g., ``place the cup on the table'')
    \item Manipulation action (e.g., ``close the drawer'', ``open the microwave'')
\end{itemize} \\

Rules & 
\mbox{}\begin{itemize}[nosep, leftmargin=*, after=\vspace{-\baselineskip}]
    \item Keep subtask descriptions short and clear.
    \item Preserve the object and location names from the original task description.
\end{itemize} \\
\\
Output Format &
\begin{minipage}[t]{\linewidth}
\ttfamily
\{ \\
\quad "subtasks": ["subtask 1", "subtask 2", ...] \\
\}
\end{minipage} \\

\bottomrule
\end{tabularx}
\end{table*}

\begin{table*}[t]
\centering
\scriptsize
\setlength{\tabcolsep}{6pt}
\renewcommand{\arraystretch}{1.5}
\caption{Prompt structure used for subtask completion detection. The VLM planner determines whether the current subtask has been completed using visual evidence and gripper state information.}
\label{tab:subtask_detection_prompt}

\begin{tabularx}{\textwidth}{lX}
\toprule
\textbf{Section} & \textbf{Prompt Content} \\
\midrule

Task & \texttt{\{task\_description\}} \\

Current Subtask & \texttt{\{current\_subtask\}} \\

Next Subtask & \texttt{\{next\_subtask\}} or ``None (this is the last subtask)'' \\

Visual Context & 
\begin{minipage}[t]{\linewidth}
\begin{itemize}[nosep, leftmargin=*]
    \item Image A: frame when the current subtask started
    \item Image B: current frame
\end{itemize}
\smallskip
Use visual differences between Image A and Image B to judge whether the current subtask has been completed.
\end{minipage} \\

Gripper State & 
\begin{minipage}[t]{\linewidth}
\ttfamily
LEFT\_GRIPPER = \{state\} \\
RIGHT\_GRIPPER = \{state\}
\end{minipage} \\

& 
\begin{minipage}[t]{\linewidth}
\smallskip
Optional notification if gripper state changes between frames: \\
\texttt{GRIPPER STATE CHANGE DETECTED:} \\
\texttt{LEFT\_GRIPPER: \{prev\_left\} $\to$ \{curr\_left\}} \\
\texttt{RIGHT\_GRIPPER: \{prev\_right\} $\to$ \{curr\_right\}} \\
\smallskip
This change may indicate that the current subtask has been completed. \\
Consider both the visual evidence and this gripper state change when deciding whether to proceed to the next subtask.
\end{minipage} \\

Decision Requirement & 
\begin{minipage}[t]{\linewidth}
Decide whether to:
\begin{itemize}[nosep, leftmargin=*]
    \item CONTINUE the current subtask "\{current\_subtask\}" (if it's still in progress)
    \item PROCEED to the next subtask (if the current subtask is completed)
\end{itemize}
Proceed only if the current subtask appears completed based on visual evidence and gripper state.
\end{minipage} \\

Output Format Rules & 
\begin{minipage}[t]{\linewidth}
\begin{itemize}[nosep, leftmargin=*]
    \item \textbf{target\_object}: noun only (e.g., bottle, drawer, cabinet)
    \item \textbf{target\_location}: noun phrase if applicable (e.g., cabinet shelf, countertop surface)
    \item PICK: specify only target\_object
    \item PLACE: specify target\_object and target\_location
    \item OTHER actions (open/close/push): specify target\_object
\end{itemize} 
\end{minipage} \\

Output Format & 
\begin{minipage}[t]{\linewidth}
\ttfamily
\{ \\
\quad "reasoning": "...", \\
\quad "decision": "continue" or "proceed", \\
\quad "target\_object": "<noun>", \\
\quad "target\_location": "<location or null>" \\
\}
\end{minipage} \\

\bottomrule
\end{tabularx}
\end{table*}
\section{VLA Experimental Demonstrations}
To demonstrate the practical capabilities of VP-VLA, we visualize execution trajectories across simulation benchmarks and real-world deployments.

\begin{figure}[t]
    \centering
    \includegraphics[width=1.\linewidth]{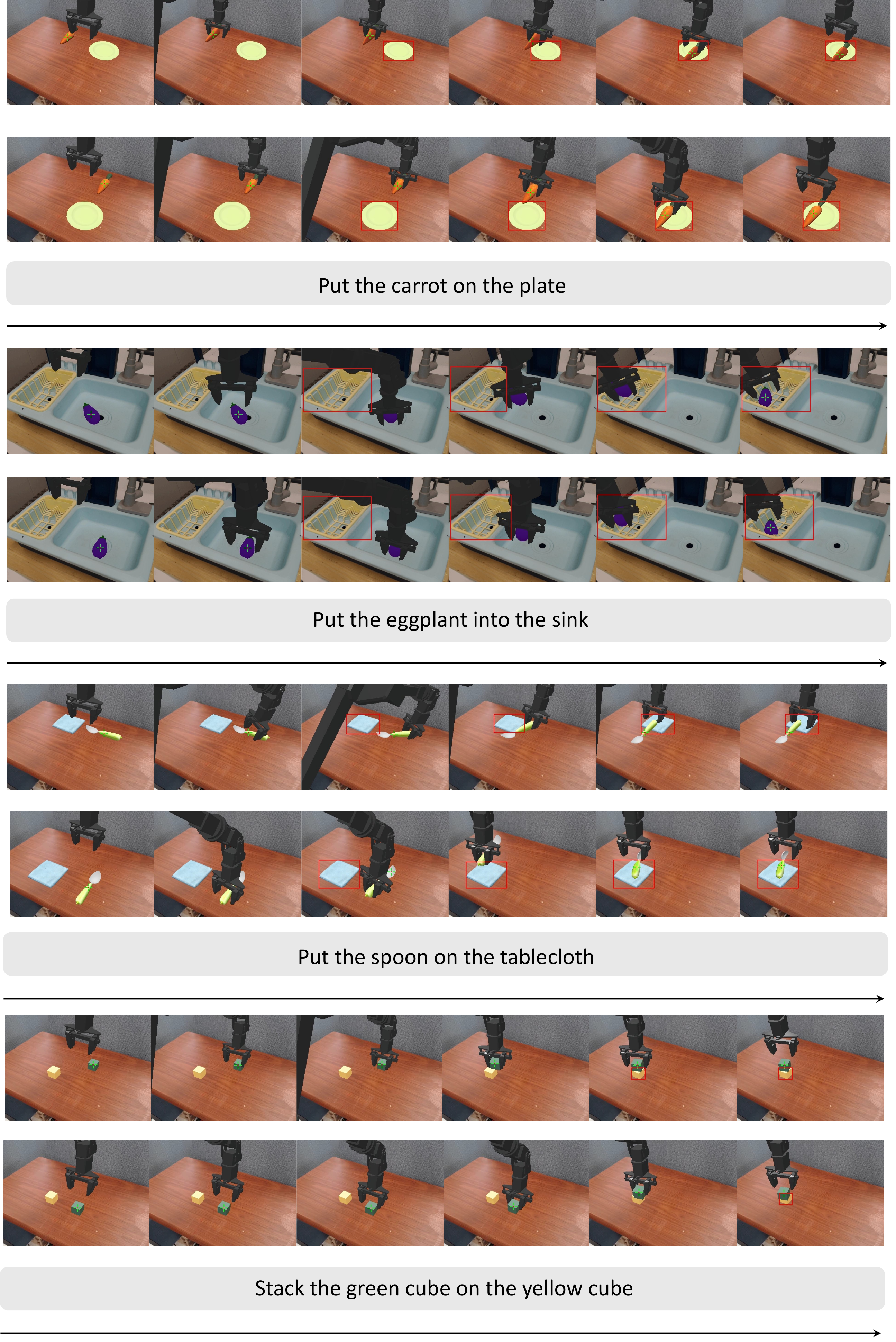}
    \caption{Inference visualization on the SimplerEnv simulation environment.}
    \label{fig:appendix_simpler}
\end{figure}

\textbf{SimplerEnv Demonstrations.} Fig.~\ref{fig:appendix_simpler} illustrates the model’s performance across four distinct tasks. Given that SimplerEnv evaluates models in out-of-distribution (OOD) settings, success requires both high-precision localization and robust control. As shown, VP-VLA accurately decompose the manipulation process into discrete stages and effectively completes each phase.

\begin{figure}[t]
    \centering
    \includegraphics[width=1.\linewidth]{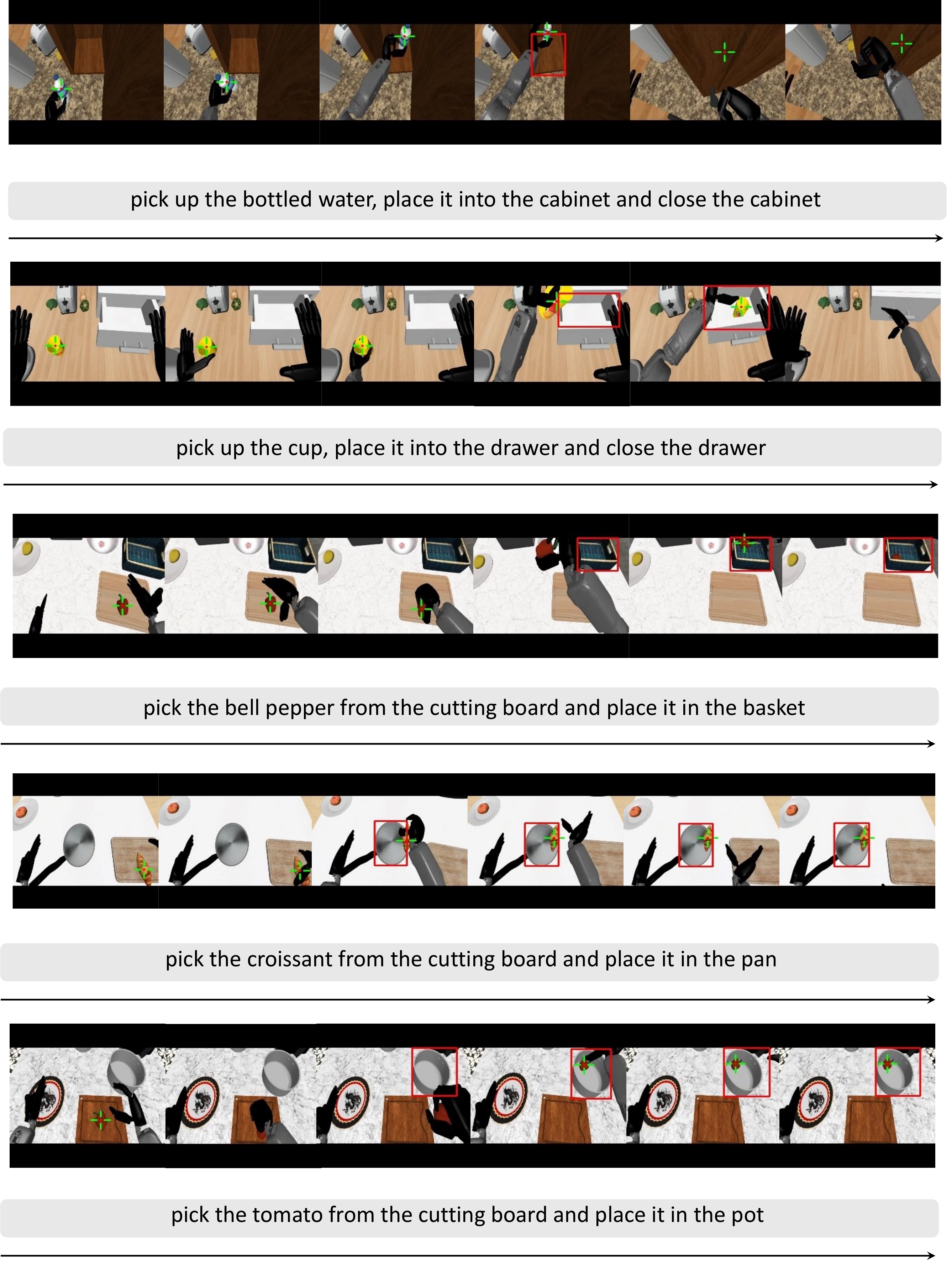}
    \caption{Inference visualization on the RoboCasa
Tabletop simulation environment.}
    \label{fig:appendix_robocasa}
\end{figure}

\textbf{RoboCasa Demonstrations.} Fig.~\ref{fig:appendix_robocasa} showcases the model's performance on the realistic RoboCasa benchmark, which involves complex, multi-step interactions in cluttered kitchen environments. These visualizations highlight VP-VLA’s ability to generalize to humanoid embodiments using egocentric observations while managing long-horizon activities.

\begin{figure}[t]
    \centering
    \includegraphics[width=1.\linewidth]{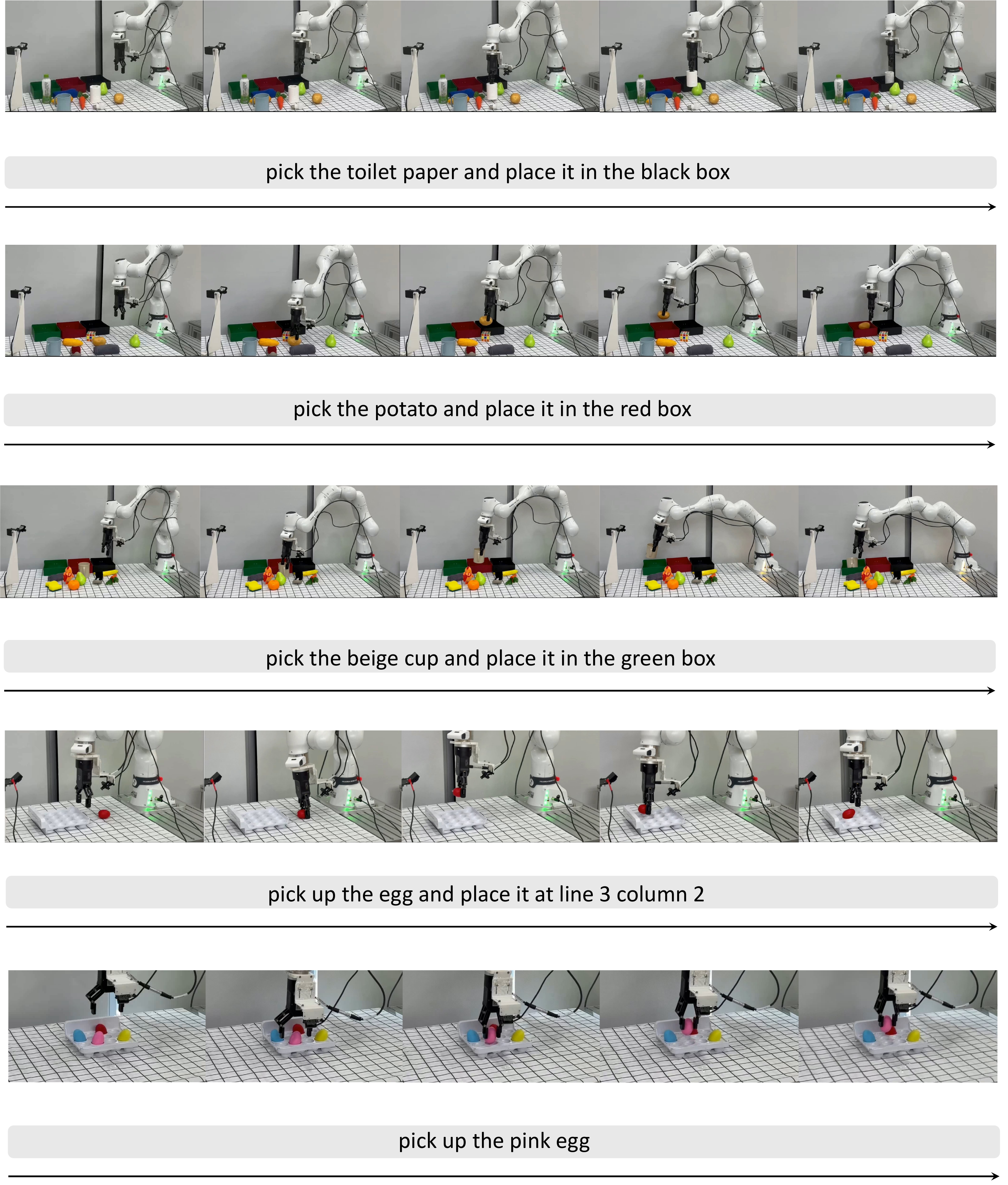}
    \caption{Inference visualization on real-world tasks.}
    \label{fig:appendix_real_world}
\end{figure}

\textbf{Real-World Demonstrations.} Fig.~\ref{fig:appendix_real_world} illustrates the VP-VLA inference process in a physical environment. To achieve a successful rollout, the model must maintain precise reasoning and grounding in novel objects and unseen placement setting. The demonstration confirms that our method facilitates smooth object manipulation and placement in both in-distribution and OOD scenarios.

\begin{table}[t]
\scriptsize
\centering
\caption{Categorization results (10k steps). Successes out of 10 trials. Bold indicates best performance.}
\label{tab:app_categorization_results}
\setlength{\tabcolsep}{4pt}
\begin{tabular}{lcc @{\hskip 0.5cm} ccc}
\toprule
\textbf{Object} & \textbf{Ours} & \textbf{QwenOFT} & \textbf{Object} & \textbf{Ours} & \textbf{QwenOFT} \\
\midrule
\multicolumn{3}{l}{\textit{Green Box}} & \multicolumn{3}{l}{\textit{Red Box}} \\
\cmidrule(r){1-3} \cmidrule(l){4-6}
Bottle (ID)      & 7  & 7 & Banana (ID) & 8  & 6 \\
Beige cup (ID)   & 9  & 8 & Orange (ID) & 9  & 9 \\
Black shoe (ID)  & 10 & 9 & Corn (ID)   & 10 & 9 \\
Toy chicken (ID) & 8  & 7 & Potato (ID) & 10 & 10 \\
Blue cup (OOD)   & 8  & 8 & Pear (OOD)  & 9  & 8 \\
Red shoe (OOD)   & 9  & 7 & Carrot (OOD)& 8  & 7 \\

\midrule

\multicolumn{3}{l}{\textit{Black Box}} & \multicolumn{3}{c}{\textbf{Summary Statistics}} \\
\cmidrule(r){1-3} \cmidrule(l){4-6}
 &  &  & \textbf{Summary} & \textbf{ID/OOD} & \textbf{ID/OOD} \\
Towel (ID)        & 10 & 10 & \textbf{Green} Avg & 34/17 & 31/15 \\
Light bulb (ID)   & 6  & 4  & \textbf{Red} Avg   & 37/17 & 34/15 \\
Rubik's cube (ID) & 10 & 9  & \textbf{Black} Avg & 34/17 & 31/8 \\
\cmidrule(lr){4-6} 
Toilet paper (ID) & 8  & 8  & \cellcolor{gray!15}\textbf{Total ID}  & \cellcolor{gray!15}\textbf{87.5\%} & \cellcolor{gray!15}80.0\% \\
Scrambled cube (OOD)   & 9  & 3  & \cellcolor{gray!15}\textbf{Total OOD} & \cellcolor{gray!15}\textbf{85.0\%} & \cellcolor{gray!15}63.3\% \\
Sponge (OOD)      & 8  & 5  & & & \\
\bottomrule
\end{tabular}
\end{table}

\begin{table}[!t]
\centering
\caption{Pick colored egg results. Performance (raw score/12) of our method vs. QwenOFT. We evaluate across In-Domain (ID) and two OOD scenarios (Color and Position change).}
\label{tab:app_colored_egg_results}
\scriptsize
\setlength{\tabcolsep}{4pt}
\begin{tabular}{l cc c cc c cc}
\toprule
& \multicolumn{2}{c}{\textbf{In-Domain}} && \multicolumn{2}{c}{\textbf{OOD (Color)}} && \multicolumn{2}{c}{\textbf{OOD (Pos.)}} \\
\cmidrule{2-3} \cmidrule{5-6} \cmidrule{8-9}
\textbf{Color} & Ours & QwenOFT & & Ours & QwenOFT & & Ours & QwenOFT \\
\midrule
Blue   & 10 & 7 &         &   &   &        &   &   \\
Pink   & 8  & 6 & Purple & 9 & 4 & Blue & 9 & 6 \\
Red    & 10 & 8 & Green  & 9 & 3 & Red  & 9 & 7   \\
Yellow & 9  & 7 &          &   &   &        &   &   \\
\midrule
\textbf{Avg (\%)} & \textbf{77.1} & 58.3 && \textbf{75.0} & 29.2 && \textbf{75.0} & 54.2 \\
\bottomrule
\end{tabular}
\end{table}

\begin{table}[!t]
\centering
\caption{Egg carton placement results. Successes are reported out of 5 trials. L$X$ C$Y$ denotes the position at Line $X$, Column $Y$. We report both In-Domain (ID) and Out-of-Distribution (OOD) performance.}
\label{tab:app_egg_grid_results}
\scriptsize
\setlength{\tabcolsep}{5pt}
\begin{tabular}{lcc @{\hskip 0.6cm} lcc}
\toprule
\multicolumn{3}{c}{\textbf{In-Domain (ID)}} & \multicolumn{3}{c}{\textbf{Out-of-Distribution (OOD)}} \\
\cmidrule(r){1-3} \cmidrule(l){4-6}
\textbf{Pos.} & \textbf{Ours} & \textbf{QwenOFT} & \textbf{Pos.} & \textbf{Ours} & \textbf{QwenOFT} \\
\midrule
L1 C2 & 5.00 & 4.25 & L1 C3 & 2.50 & 1.75 \\
L1 C4 & 5.00 & 4.50 & L2 C1 & 2.25 & 2.50 \\
L2 C4 & 5.00 & 2.25 & L3 C2 & 5.00 & 3.25 \\
L2 C2 & 4.00 & 4.25 & L4 C3 & 4.00 & 3.50 \\
\cmidrule(l){4-6} 
L3 C1 & 4.50 & 4.50 & \multicolumn{3}{l}{\textbf{Summary Statistics}} \\
\cmidrule(l){4-6}
L3 C3 & 4.50 & 1.00 & \textbf{Group} & \textbf{Ours} & \textbf{QwenOFT} \\
\cmidrule(l){4-6}
L4 C2 & 4.50 & 3.00 & \cellcolor{gray!15}ID Avg. & \cellcolor{gray!15}\textbf{91.3\%} & \cellcolor{gray!15}70.6\% \\
L4 C4 & 4.00 & 4.50 & \cellcolor{gray!15}OOD Avg. & \cellcolor{gray!15}\textbf{68.8\%} & \cellcolor{gray!15}55.0\% \\
\bottomrule
\end{tabular}
\end{table}

\begin{table*}[!t]
    \centering
    \small
    \renewcommand{\arraystretch}{1.4} 
    \setlength{\tabcolsep}{1.6pt} 

    \caption{
      Results of evaluating the VLA models with the GR1 robot in the RoboCasa Tabletop simulation environment. We highlight the best result in ~\textbf{bold} and the second-best results with \underline{underline}.
    }
     \begin{adjustbox}{width=\linewidth}
    \begin{tabular}{l c c c c c c c}
        \toprule
        \rowcolor{white} 
        {Task} & 
        {\scriptsize \makecell{\textbf{Isaac-GR00T}\\\textbf{N1.5}}} & 
        {\scriptsize \makecell{\textbf{Isaac-GR00T}\\\textbf{N1.6}}} & 
        {\scriptsize \makecell{\textbf{QwenGR00T}\\\textbf{+Qwen3VL}}} & 
        {\scriptsize \makecell{\textbf{QwenPI}\\\textbf{+Qwen3VL}}} & 
        {\scriptsize \makecell{\textbf{QwenOFT}\\\textbf{+Qwen3VL}}} & 
        {\scriptsize \makecell{\textbf{QwenFAST}\\\textbf{+Qwen3VL}}} &
        {\scriptsize \makecell{\textbf{Ours}\\\textbf{+Qwen3VL}}}\\
        \midrule
        PnP Bottle To Cabinet Close                                    & 54.0 & 51.5 & 46.0 & 26.0 & 30.0 & 38.0 & 54.0\\
        PnP Can To Drawer Close                                        & 50.0 & 13.0 & 80.0 & 62.0 & 76.0 & 44.0 & 72.0\\
        PnP Cup To Drawer Close                                        & 38.0 &  8.5 & 54.0 & 42.0 & 44.0 & 56.0 & 44.0\\
        PnP Milk To Microwave Close                                    & 60.0 & 14.0 & 48.0 & 50.0 & 44.0 & 44.0 & 74.0\\
        PnP Potato To Microwave Close                                  & 32.0 & 41.5 & 28.0 & 42.0 & 32.0 & 14.0 & 34.0\\
        PnP Wine To Cabinet Close                                      & 38.0 & 16.5 & 46.0 & 32.0 & 36.0 & 14.0 & 48.0\\
        \midrule
        \rowcolor{gray!20}\textbf{PnP * to * Close (Avg)}                       & 45.3 & 24.2 & 50.3 & 42.3 & 43.7 & 35.0 & 54.3\\
        \midrule
        PnP Novel From Cuttingboard To Basket                          & 38.0 & 58.0 & 48.0 & 40.0 & 50.0 & 54.0 & 66.0\\
        PnP Novel From Cuttingboard To Cardboardbox                    & 46.0 & 46.5 & 40.0 & 46.0 & 40.0 & 42.0 & 54.0\\
        PnP Novel From Cuttingboard To Pan                             & 58.0 & 68.5 & 68.0 & 60.0 & 70.0 & 58.0 & 74.0\\
        PnP Novel From Cuttingboard To Pot                             & 62.0 & 65.0 & 52.0 & 40.0 & 54.0 & 58.0 & 54.0\\
        PnP Novel From Cuttingboard To Tieredbasket                    & 28.0 & 46.5 & 56.0 & 44.0 & 38.0 & 40.0 & 56.0\\
        \midrule
        \rowcolor{gray!20}\textbf{PnP Novel From Cuttingboard To * (Avg)}       & 46.4 & 56.9 & 52.8 & 46.0 & 50.4 & 50.4 & 60.8\\
        \midrule
        PnP Novel From Placemat To Basket                              & 30.0 & 58.5 & 42.0 & 44.0 & 32.0 & 36.0 & 48.0 \\
        PnP Novel From Placemat To Bowl                                & 60.0 & 57.5 & 44.0 & 52.0 & 58.0 & 38.0 & 74.0\\
        PnP Novel From Placemat To Plate                               & 56.0 & 63.0 & 48.0 & 50.0 & 52.0 & 42.0 & 70.0\\
        PnP Novel From Placemat To Tieredshelf                         & 36.0 & 28.5 & 18.0 & 28.0 & 24.0 & 18.0 & 26.0\\
        \midrule
        \rowcolor{gray!20}\textbf{PnP Novel From Placemat To * (Avg)}           & 45.5 & 51.9 & 38.0 & 43.5 & 41.5 & 33.5 & 54.5\\
        \midrule
        PnP Novel From Tray To Cardboardbox                            & 52.0 & 51.5 & 38.0 & 34.0 & 44.0 & 28.0 & 44.0 \\
        PnP Novel From Tray To Plate                                   & 48.0 & 71.0 & 56.0 & 64.0 & 56.0 & 34.0 & 66.0\\
        PnP Novel From Tray To Pot                                     & 60.0 & 64.5 & 50.0 & 44.0 & 62.0 & 46.0 & 38.0\\
        PnP Novel From Tray To Tieredbasket                            & 52.0 & 57.0 & 36.0 & 50.0 & 54.0 & 36.0 & 58.0\\
        PnP Novel From Tray To Tieredshelf                             & 32.0 & 31.5 & 16.0 & 28.0 & 30.0 & 16.0 & 24.0\\
        \midrule
        \rowcolor{gray!20}\textbf{PnP Novel From Tray To * (Avg)}               & 48.8 & 55.1 & 39.2 & 44.0 & 49.2 & 32.0 & 46.0\\
        \midrule
        PnP Novel From Plate To Bowl                                   & 58.0 & 57.0 & 60.0 & 52.0 & 60.0 & 52.0 & 52.0 \\
        PnP Novel From Plate To Cardboardbox                           & 44.0 & 43.5 & 50.0 & 40.0 & 50.0 & 30.0 & 44.0 \\
        PnP Novel From Plate To Pan                                    & 60.0 & 51.0 & 54.0 & 36.0 & 66.0 & 48.0 & 56.0\\
        PnP Novel From Plate To Plate                                  & 64.0 & 78.7 & 70.0 & 48.0 & 68.0 & 50.0 & 62.0\\
        \midrule
        \rowcolor{gray!20}\textbf{PnP Novel From Plate To * (Avg)}              & 56.5 & 57.6 & 58.5 & 44.0 & 61.0 & 45.0 &  53.5\\
        \midrule
        \rowcolor{gray!30} 
        \textbf{Average}                                                      & 48.2 & 47.6 & 47.8 & 43.9 & \underline{48.8} & 39.0 & \textbf{53.8}\\
        \bottomrule
    \end{tabular}
        \end{adjustbox}
    \label{tab:app_robocasa_main_tab}
\end{table*}


\end{document}